\documentclass[11pt]{article}

\usepackage[final]{acl}

\usepackage{times}
\usepackage{latexsym}

\usepackage[T1]{fontenc}

\usepackage[utf8]{inputenc}

\usepackage{microtype}

\usepackage{inconsolata}
\usepackage{graphicx}
\usepackage{amssymb}
\usepackage{amsmath}
\usepackage{booktabs}
\usepackage{multirow}
\usepackage{array}
\usepackage{balance}
\usepackage{tikz}
\usepackage{pgfplots}
\pgfplotsset{compat=1.18}

\usepackage{xcolor}
\usepackage{tikz}
\usetikzlibrary{positioning,calc,fit,backgrounds,arrows.meta,decorations.pathreplacing}

\usepackage{float}

\newcommand{\diffie}{\textsc{DiffIE}~}

\title{DiffIE: Diffusion-based Open Information Extraction}

\author{
  Konstantin Fedorov\textsuperscript{1,2} \qquad
  Valentin Malykh\textsuperscript{3,4,5} \\
  \textsuperscript{1}Matrosov Institute for System Dynamics and Control Theory, SB RAS,\\
  \textsuperscript{2}AI Talent Hub, ITMO University,
  \textsuperscript{3}MWS AI, \\
  \textsuperscript{4}Trusted AI Research Center, RAS,
  \textsuperscript{5}IITU University \\
  Correspondence: \texttt{k.fedorov@innopolis.university}
}

\begin{document}
\maketitle
\begin{abstract}
A single sentence often expresses multiple valid relational triplets,
which makes Open Information Extraction (OpenIE) fundamentally a
multi-output task. Existing neural systems handle this by autoregressive
generation, which is flexible but slow and prone to redundancy, or by
fixed-slot prediction, which is efficient but couples the extraction
budget to training. We introduce \diffie which instead treats the
stochasticity of conditional discrete diffusion as the extraction
mechanism itself: independent reverse-diffusion trajectories over
per-token role tags produce a pool of candidate triplets, which are
clustered under lenient matching and ranked to form the output. Both
the pool size and the number of returned extractions are inference-time
choices, decoupling the extraction budget from training and exposing
test-time compute as a tunable axis.
\diffie achieves the new state of the art in CaRB~(1-1) both F1 and AUC, and outperforms the strongest rule-based system (ClausIE) in BenchIE; it also remains competitive in standard CaRB and WiRe57 evaluations, giving the best average score among systems that report all four benchmarks. Ablations
show that uniform discrete diffusion outperforms
absorbing-state diffusion in our setting, and that a matched
non-diffusion stochastic tagger does not reproduce its gains. Our
results indicate that diffusion stochasticity is an effective mechanism
for structured prediction tasks with multiple valid outputs.
\end{abstract}

\begin{figure}[tbh!]
\centering
\resizebox{0.8\linewidth}{!}{%
    \input{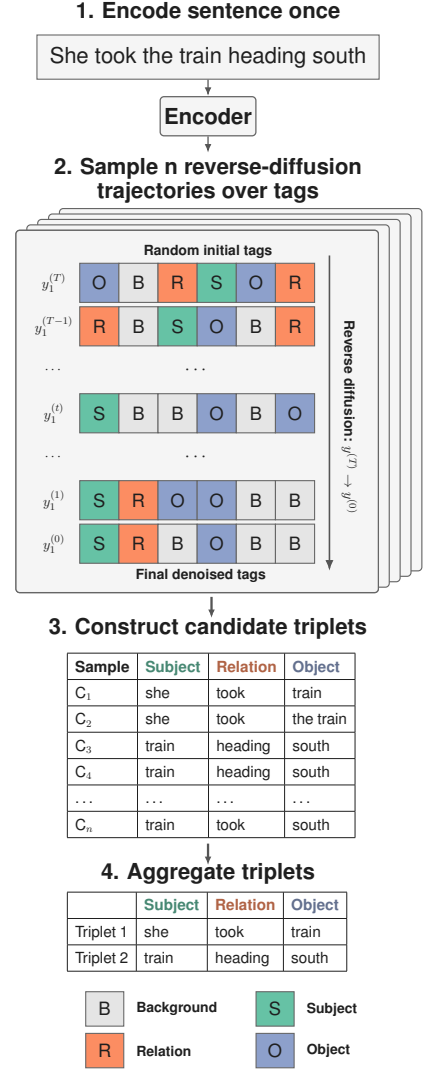}
}
\caption{Overview of \diffie inference. Multiple reverse-diffusion trajectories conditioned on the same sentence produce candidate triplets, which may include duplicates and noisy outputs. Lenient aggregation clusters candidates and returns the top-ranked extractions.}
\label{fig:diffie-overivew}
\end{figure}
\section{Introduction}

Open Information Extraction (OpenIE) aims to extract schema-free relational triplets (Subject, Relation, Object) from natural language text, where a single sentence typically expresses multiple distinct facts. Existing neural approaches divide into autoregressive sequence generators \citep{cui-etal-2018-neural,kolluru-etal-2020-imojie,chen2024dualoie}, which are flexible but slow and prone to redundancy, and sequence-labeling models \citep{stanovsky-etal-2018-supervised,kolluru-etal-2020-openie6,zhan2020span}, which are efficient but typically produce one triplet per pass. DetIE \citep{Vasilkovsky2022detie} addresses this with an object-detection-inspired design that predicts a fixed number $N$ of candidates in a single forward pass, demonstrating that non-autoregressive multi-triplet extraction is viable; the trade-off is that $N$ is fixed at training time and the model must be retrained when the target cardinality changes.

We propose \textsc{DiffIE}, a non-autoregressive sequence-labeling model based on conditional discrete diffusion over per-token role tags. Rather than decoding a single tag sequence, \diffie exploits the stochasticity of the reverse diffusion process: $n$ independent denoising trajectories yield a sample-based candidate pool of valid tag sequences, which a lenient-match extractor aggregates into a final triplet set. Because $n$ is set at inference time, it doubles as a tunable compute--quality axis: more samples expand the candidate pool at greater inference cost. The size of the returned extraction set is likewise an inference-time choice, so the extraction budget can be retuned per corpus without retraining.

Our contributions are: (1) the first application of discrete diffusion language models to OpenIE, formulating it as conditional sequence labeling; (2) a sample-aggregation inference mechanism, paired with a lenient-match extractor tailored to the multi-reference nature of OpenIE ground truth, that decouples the extraction budget from training; (3) an empirical demonstration that uniform-noise discrete diffusion (D3PM) outperforms absorbing-state diffusion (MDLM) in the extreme small-vocabulary regime ($|\mathcal{V}|=4$); (4) a matched non-diffusion stochastic tagger control that isolates reverse diffusion, rather than repeated sampling and clustering, as the source of the gains; and (5) best reported CaRB (1-1) F1 and AUC and BenchIE F1, outperforming prior neural systems and the strongest rule-based baseline on BenchIE.\footnote{Code and checkpoints: \url{https://github.com/KonstFed/DiffIE}.}

\section{Related Work}
\label{sec:related}
    \subsection{Neural Open Information Extraction}
    \label{sec:related:neural-openie}

    \paragraph{Sequence-labeling approaches}
    treat extraction as token-level tagging, typically with BIO-style schemes \citep{stanovsky-etal-2018-supervised}. OpenIE6 \citep{kolluru-etal-2020-openie6} replaces flat tagging with a 2-D Iterative Grid Labeling formulation that captures discontinuous spans and overlapping relations, and SpanOIE \citep{zhan2020span} first detects predicate spans and then classifies their arguments. Most relevant to our work, DetIE \citep{Vasilkovsky2022detie} adopts an object-detection-inspired formulation that emits a fixed number $N$ of triplet candidates in a single forward pass with bipartite matching at training time. Our method belongs to this family but replaces the train-time fixed budget with inference-time sampling from a distributional model (\S\ref{sec:method}).

    \paragraph{Sequence-generation approaches}
    formulate OpenIE as autoregressive text-to-text generation. Early work introduced encoder--decoder copy mechanisms \citep{cui-etal-2018-neural}; IMoJIE \citep{kolluru-etal-2020-imojie} produces extractions iteratively, conditioning each on previously generated ones. More recent systems include T5-based formulations \citep{fan-he-2023-efficient} and dual learning to reduce missing and redundant triples \citep{chen2024dualoie}. CycleOIE \citep{cycleoie2025} introduces a low-resource training framework whose curated \textsc{LSOIE-examples} subset we use as our training corpus. Generative approaches are flexible but pay an autoregressive inference cost and tend to produce redundant extractions, motivating non-autoregressive alternatives like~\citep{Vasilkovsky2022detie}, of which ours is one.

    \paragraph{Low-resource OpenIE.}
    CycleOIE~\citep{cycleoie2025} argues that neural OpenIE remains
    heavily dependent on large annotated corpora and curates small,
    GPT-annotated training sets via two prompting strategies, reporting
    stronger results from the few-shot examples variant. We use this
    \textsc{lsoie-examples} subset as our training source and later
    analyze whether augmenting it with raw LSOIE helps.

    \paragraph{Relation to LLM prompting.}
    Recent work prompts LLMs to perform OpenIE directly
    \citep{chen2024dualoie}; such systems are autoregressive, expensive at
    corpus scale, and offer little control over the extraction
    distribution. \diffie encodes each sentence once and exposes test-time
    compute as a tunable axis, so its cost can be set after training rather
    than fixed by a decoder (\S\ref{sec:ablations-nstudy}).

    \subsection{Discrete Diffusion Language Models}
    \label{sec:related:discrete-diffusion}

    Diffusion models for text fall into continuous-space variants---which embed discrete tokens into a continuous latent and diffuse there \citep{li2022diffusionlm}---and discrete-space variants, which define the noising process directly over the categorical vocabulary. D3PM \citep{austin2021d3pm} establishes the discrete-diffusion framework with several transition matrix choices, of which absorbing-state and uniform corruption are the two most widely used. MDLM \citep{sahoo2024simple} simplifies absorbing-state diffusion and, alongside SEDD \citep{lou2024sedd} and LLaDA \citep{nie2025llada}, has established discrete diffusion as a viable alternative to autoregressive language modeling.

    The relative performance of the two formulations depends on vocabulary size: \citet{schiff2025simple} show that uniform-noise diffusion can match or exceed absorbing-state on small-vocabulary language modeling, contrary to the common assumption that absorbing-state is uniformly stronger. Our four-symbol tag vocabulary is an extreme instance of this regime; we implement both and report the comparison in \S\ref{sec:ablations-mdlm}.

 \subsection{Diffusion for Structured Prediction}
    \label{sec:related:diffusion-for-tagging}
    Prior diffusion-based approaches to structured NLP either operate in
    continuous space (DiffusionNER \citep{diffusionner}) or assume a fixed
    relation schema (IPED \citep{zhao-iped}), making them incompatible
    with schema-free, multi-extraction OpenIE. Closest to our work,
    DiffusionSL \citep{diffusionsl} performs sequence labeling via a
    \emph{Bit-Tag Converter} that encodes each tag as a bit pattern and
    applies continuous Gaussian diffusion in bit space; it is evaluated
    on tasks (NER, POS, chunking) where each sentence has a single best
    label sequence and decodes from a single trajectory. \diffie differs
    in three ways: we use categorical discrete diffusion over the
    four-symbol tag vocabulary directly; we draw $n$ trajectories at
    inference and aggregate them, treating sample diversity as the
    mechanism for capturing multiple valid extractions; and we introduce
    task-specific extractors (\S\ref{sec:method:inference}) for the
    multi-reference nature of OpenIE ground truth.

    \subsection{OpenIE Evaluation Benchmarks}
    \label{sec:related:benchmarks}
    OpenIE evaluation is complicated by the fact that a single sentence
    admits many valid extractions with substantial surface-form variation,
    and benchmarks differ in both reference construction and matching.

    \paragraph{CaRB and CaRB (1-1).}
    CaRB \citep{bhardwaj-etal-2019-carb} provides crowdsourced extractions
    for 1{,}282 sentences with a token-level matcher allowing partial
    credit. The standard matcher permits many-to-one alignment scored by
    token overlap, which rewards systems that emit overly long extractions
    covering many gold tokens at once
    \citep{lechelle-etal-2019-wire57,gashteovski-etal-2022-benchie,fatahi-bayat-etal-2022-compactie}.
    CaRB~(1-1) enforces one-to-one alignment via the Hungarian algorithm
    and is widely regarded as a more faithful measure of extraction
    quality.

    \paragraph{BenchIE.}
    BenchIE \citep{gashteovski-etal-2022-benchie} replaces per-triplet
    gold annotations with \emph{fact synsets}---exhaustive clusters of
    acceptable surface realizations of the same underlying fact---and
    applies strict synset-level matching. Because credit requires hitting
    a fact rather than fragments of one, BenchIE is substantially harder
    to game by over-extraction.

    \paragraph{WiRe57.}
    WiRe57 \citep{lechelle-etal-2019-wire57} is a smaller benchmark
    (57 sentences) with manually curated, high-precision references,
    providing a complementary sanity check.

    \paragraph{Benchmark selection.}
    We evaluate \diffie on all four, treating CaRB~(1-1) and BenchIE as
    the more rigorous indicators of extraction quality. The suite spans
    the full lenient-to-strict matching spectrum, letting us characterize
    where our method's strengths lie.

\section{Method}
\label{sec:method}

    We formulate OpenIE as conditional discrete diffusion over per token tag sequences. Given an input sentence, our model learns to reverse a discrete corruption process that maps a random tag sequence to the ground-truth tagging. At inference time, we exploit the stochasticity of this reverse process: by drawing many independent denoising trajectories from the same sentence, we obtain a sample-based candidate pool of valid tag sequences, which we then aggregate into a final set of (Subject, Relation, Object) triplets.

    \subsection{Problem Formulation}
    \label{sec:method:formulation}

    Let $x = (x_1, \ldots, x_L)$ denote an input sentence of $L$ tokens. Following prior sequence-labeling formulations of OpenIE \citep{Vasilkovsky2022detie,kolluru-etal-2020-openie6}, we represent extractions as a tag sequence $y = (y_1, \ldots, y_L)$ where each $y_i \in \mathcal{V} = \{B, S, R, O\}$ assigns a role---Background, Subject, Relation, or Object---to the $i$-th token. A single tag sequence encodes exactly one triplet, and in the current post-sampling construction step we recover the Subject, Relation, and Object spans as the longest contiguous runs of their respective tags.

    Unlike DetIE \citep{Vasilkovsky2022detie}, which predicts a fixed number $N$ of triplets simultaneously and resolves multi-extraction via bipartite matching at train time, and unlike IMoJIE \citep{kolluru-etal-2020-imojie}, which generates triplets autoregressively, our formulation produces \emph{one tag sequence per forward sample}. Multi-extraction is recovered at inference time through repeated stochastic sampling (\S\ref{sec:method:inference}).

    \subsection{Architecture}
    \label{sec:method:arch}

    \diffie consists of two components: a pretrained transformer \textit{encoder} and a small randomly-initialized diffusion \textit{denoiser}. An overview is shown in Figure~\ref{fig:diffie-overivew}.

    \paragraph{Encoder.}
    We use a pretrained transformer encoder $\mathrm{Enc}(\cdot)$ to map the input sentence to contextual token embeddings $h^{\text{enc}} \in \mathbb{R}^{L \times d}$. Depending on the tuned configuration, some lower encoder layers may be frozen while the remaining layers are fine-tuned jointly with the denoiser.

    \paragraph{Denoiser.}
    The denoiser is a small transformer with self-attention only and is trained from scratch. At denoising step $t$, it takes the current noisy tag sequence $y^{(t)}$, embeds it as $h^{\text{tag}} \in \mathbb{R}^{L \times d}$, and projects the encoder context to the same dimension. Conditioning is implemented by tokenwise fusion: for each position $i$, the tag embedding and contextual embedding are concatenated and mapped through a learned projection to obtain a shared latent representation. The resulting sequence is then processed exclusively by self-attention, so the encoder information enters the denoiser without a separate cross-attention module. A timestep embedding is added before the self-attention stack, and the decoder outputs logits over the tag states for each token position.

    \subsection{Discrete Diffusion Training}
    \label{sec:method:training}

    A sentence with $m$ gold triplets is expanded into $m$ independent (sentence, tag-sequence) examples, one per triplet, which are then shuffled into mini-batches. The denoiser thus learns the marginal distribution over single-triplet taggings conditioned on the sentence; this marginal is sampled repeatedly at inference to recover the variable-cardinality extraction set.

    We adopt the uniform discrete diffusion formulation of \citet{austin2021d3pm}. The forward process gradually corrupts the ground-truth tag sequence $y^{(0)} = y^{*}$ by transitioning each token toward a uniform draw from $\mathcal{V}$ according to a noise schedule $\bar{\alpha}_t$:
    \begin{equation}
    \label{eq:forward}
    q(y^{(t)}_i \mid y^{(0)}_i) = \bar{\alpha}_t \, \mathbf{e}_{y^{(0)}_i} + (1 - \bar{\alpha}_t) \, \tfrac{1}{|\mathcal{V}|} \mathbf{1},
    \end{equation}
    where $\mathbf{e}_{y^{(0)}_i}$ is the one-hot vector at position $y^{(0)}_i$ and $\mathbf{1}$ is the all-ones vector over $\mathcal{V}$.

    The denoising model $p_\theta(y^{(0)} \mid y^{(t)}, x)$ is trained with the standard D3PM uniform-kernel denoising objective, implemented as token-level cross-entropy at a uniformly sampled timestep $t$. To mitigate label imbalance, we use per-class loss weights during training.

    \subsection{Sample-Aggregation Inference}
    \label{sec:method:inference}

    Because OpenIE admits multiple valid extractions per sentence, single-trajectory decoding is fundamentally limited. We instead exploit the stochasticity of the reverse diffusion process to build a sample-based candidate pool.

    \paragraph{Sampling.}
    Given a sentence $x$, we encode it once and run $n$ independent reverse-diffusion trajectories, each initialized from a uniform random tag sequence $y^{(T)} \sim \mathrm{Unif}(\mathcal{V}^L)$. In implementation, the $n$ noisy tag sequences are stacked along the batch dimension and denoised in parallel for $T=16$ steps, reusing the same encoder states. Each trajectory yields a denoised tag sequence $y^{(0)}_k$ for $k = 1, \ldots, n$.

    \paragraph{Triplet construction.}
    The diffusion model predicts token-level role tags and does not require role labels to be contiguous. In the current post-sampling construction step, we surface the longest contiguous span for each role. This simple heuristic filters isolated tag errors and works well in our experiments, but it may drop useful tokens when a predicted role is discontinuous. Because this step is applied only after sampling, alternative triplet-construction rules can be used without retraining the model. Future work could replace the longest-span heuristic with construction rules that preserve multiple predicted spans per role. If any of the three role tags is absent from $y^{(0)}_k$, the sample produces no triplet.

    \paragraph{Aggregation.}
    The $n$ trajectories yield candidate triplets $C_1, \ldots, C_n$, treated
    as i.i.d.\ samples from $p_\theta(\cdot \mid x)$. For any triplet $T$, the
    empirical frequency

    \begin{equation}
        \hat{p}_n(T \mid x)
        =
        \frac{1}{n}
        \sum_{k=1}^{n}
        \mathbf{1}\{C_k = T\}
        \label{eq:freq_score}
    \end{equation}
    is the empirical frequency of each candidate or candidate cluster. We use
    these frequencies as confidence scores in two extractors.

    \paragraph{Lenient-match extractor.}
    To recover this dispersed mass, we cluster candidates under CaRB-style
    lenient matching \citep{bhardwaj-etal-2019-carb}. For triplets $T^{(i)}$
    and $T^{(j)}$ with total word counts $n_i, n_j$, let $m_{ij}$ be the
    role-wise lowercased word-multiset overlap, summed over subject, relation,
    and object. The symmetric lenient F1 is
    \begin{equation}
        F_{ij} \;=\; \frac{2\,m_{ij}}{n_i + n_j},
        \label{eq:lenient_f1}
    \end{equation}
    matching the CaRB evaluator's token-level matcher. Triplets with
    $F_{ij} \geq \tau$ are declared equivalent, and we take connected components
    under this relation as clusters. Each cluster's mass is the sum of its
    members' frequencies; we return the top-$k$ clusters by mass, each surfaced
    by its highest-frequency member. As $\tau \to 1$ the extractor reduces to
    the frequency baseline; $\tau$, $k$, and $n$ are tuned on validation data.
    We fix the output budget to $k=4$ and the lenient-clustering threshold to
    $\tau=0.9$, selected on the CaRB development set. The sample count $n$
    controls the size of the candidate pool, while $k$ controls how many
    clustered extractions are returned. Unlike fixed-slot models, changing $k$
    at inference time does not require retraining.

    \paragraph{Test-time compute control.}
    The number of samples $n$ is a hyperparameter of inference, not training. This makes $n$ a tunable compute-quality axis: larger $n$ provides a larger candidate pool at greater inference cost. We analyze this tradeoff in \S\ref{sec:ablations-nstudy}.

    \begin{table}[t]
    \centering
    \small
    \begin{tabular}{lrrr}
    \toprule
    Configuration & Sentences & Instances & Filtered \\
    \midrule
    \textsc{lsoie-ex-50}    &     50 &      267 &      171 \\
    \textsc{lsoie-ex-250}   &    250 &    1{,}313 &      781 \\
    \textsc{lsoie-ex-1.25k} &  1{,}250 &    6{,}760 &    4{,}129 \\
    \textsc{lsoie-ex-2.5k}  &  2{,}500 &   13{,}415 &    8{,}266 \\
    \textsc{lsoie-ex-full}  &  4{,}901 &   26{,}349 &   16{,}359 \\
    \midrule
    \quad + \textsc{lsoie-20k}   & 24{,}901 &   46{,}349 &   36{,}359 \\
    \quad + \textsc{lsoie-full}  & 50{,}917 &   72{,}365 &   62{,}375 \\
    \bottomrule
    \end{tabular}
    \caption{Dataset statistics for all training configurations.
      \emph{Instances}: triplet pairs after sentence decomposition.
      \emph{Filtered}: instances surviving span-alignment filtering.
      The upper block contains subsets of LSOIE-examples; the lower block
      augments \textsc{lsoie-ex-full} with raw LSOIE data (100\% retention,
      as LSOIE is already in token-labeled format).}
    \label{tab:data_stats}
    \end{table}

    \begin{table*}[t]
    \centering
    \resizebox{\textwidth}{!}{
    \begin{tabular}{l| cc cc cc |c}
    \toprule
    & \multicolumn{2}{c}{CaRB} & \multicolumn{2}{c}{CaRB (1-1)} & BenchIE & WiRe57 & \multirow{2}{*}{Avg} \\
    \cmidrule(lr){2-3} \cmidrule(lr){4-5} \cmidrule(lr){6-6} \cmidrule(lr){7-7}
    System & F1 & AUC & F1 & AUC & F1 & F1 \\
    \midrule
    OpenIE6 \citep{kolluru-etal-2020-openie6}            & 52.7 & 33.7 & 46.4 & 26.8 & 25.0 & \textbf{40.0} & 41.0 \\
    IMoJIE \citep{kolluru-etal-2020-imojie}              & 53.5 & 33.3 & 41.4 & 22.2 & 18.6 & 36.0 & 37.4 \\
    DetIE \citep{Vasilkovsky2022detie}                   & 52.1 & 36.7 & 40.1 & 29.3 & --- & 36.0 & --- \\
    ClausIE \citep{clausie}                              & 45.0 & 22.0 & 40.2 & 17.7 & 34.0 & 33.2 & 38.1 \\

    CompactIE \citep{fatahi-bayat-etal-2022-compactie}   & 45.0 & --- & --- & --- & 26.2 & 31.8 & --- \\
    DualOIE \citep{chen2024dualoie}                      & \textbf{56.3} & --- & 51.5 & --- & --- & --- & --- \\
    ChatGPT (Chain-of-Thought) \citep{chen2024dualoie}                & 53.4 & --- & --- & --- & --- & --- & --- \\
    CycleOIE \citep{cycleoie2025}                        & 51.2 & \textbf{39.0} & 47.4 & 33.6 & --- & --- & --- \\

    \midrule

    \diffie & 52.2 & 37.1  & \textbf{51.9} & \textbf{34.5} & \textbf{34.3} & 36.1 & \textbf{43.6} \\

    \bottomrule
    \end{tabular}
    }
    \caption{
      Main results across four OpenIE benchmarks. We compare to the strongest
      published numbers available for each benchmark. Since prior systems do not
      all report the same metrics, we emphasize per-benchmark comparisons rather
      than a single aggregate score; \emph{Avg} is defined only for systems
      reporting all four benchmarks. \diffie results are means over 10 random seeds.
      BenchIE and WiRe57 scores for OpenIE6 are from \citet{cycleoie2025};
      BenchIE scores for IMoJIE and ClausIE are from \citet{fatahi-bayat-etal-2022-compactie}
      and \citet{gashteovski-etal-2022-benchie} respectively; remaining baseline scores are from their
      original papers. \textbf{Bold}: best in column.
    }
    \label{tab:main_results}
    \end{table*}

\section{Experimental Setup}
\label{sec:experimental-setup}

    \paragraph{Training data.}
    We train on LSOIE-examples~\citep{cycleoie2025}, a 4{,}901-sentence subset of LSOIE~\citep{solawetz-larson-2021-lsoie} curated by CycleOIE via example-guided prompting. We use this variant because CycleOIE reports stronger performance for example-guided curation than for its principles-guided alternative, particularly in recall and F1. Since LSOIE-examples is distributed in a generative format pairing each sentence with one or more extracted triplets, we convert it to sequence-labeling supervision by decomposing each sentence into one (sentence, triplet) instance per triplet and assigning per-token B/S/R/O labels via span alignment to tokenizer offsets. Alignment is filtered at the triplet level: instances for which the alignment fails to cover all tokens of any argument span are discarded, while the sentence is retained if at least one of its triplets aligns successfully. This filtering removes 38--40\% of instances across configurations, a stable rate attributable to CycleOIE's tendency to paraphrase argument spans rather than copy them verbatim from the source sentence. We study five curated subsets of varying size (50--4{,}901 sentences) and two configurations that augment the full curated set with raw LSOIE data. Statistics for all configurations are reported in Table~\ref{tab:data_stats}. We adopt \textsc{lsoie-ex-2.5k} as our primary configuration; the ablation justifying this choice is presented in \S\ref{sec:ablations-data}.

    \paragraph{Implementation details.}
    We use \texttt{bert-base-uncased} as the encoder, with the four lowest
    transformer layers frozen during training. The denoiser is a 6-layer
    self-attention transformer with model dimension 512, 8 attention heads,
    and concatenation-based encoder fusion. The diffusion process uses $T=16$
    steps with a cosine noise schedule ($s=0.002$). We train with AdamW,
    using a learning rate of $2\times10^{-4}$ for the denoiser and
    $5\times10^{-5}$ for the encoder, weight decay $0.01$, 500 linear warmup
    steps, and a batch size of 32 for 12 epochs. To mitigate label imbalance
    we assign a class weight of 0.7 to the background label and 1.0 to all
    other tags. Hyperparameters were selected by grid search on the CaRB
    development set.

    \paragraph{Evaluation protocol.}
    Unless otherwise stated, we use \textsc{lsoie-ex-2.5k} as the primary training configuration; \S\ref{sec:ablations-data} analyzes this choice. All \diffie results are averaged over 10 independent random seeds. The denoising sample count $n$, output budget $k$, and lenient-matching threshold $\tau$ are fixed globally using the CaRB development set.

\section{Results}
\label{sec:results}

    \diffie achieves the best reported CaRB (1-1) F1 and AUC among published OpenIE systems. This improvement is consistent across runs: every seed exceeds the previous best reported CaRB (1-1) F1 score of 51.5. Across benchmarks, F1 variance is low, with standard deviation at most 0.5 points.
    Holm-corrected one-sample $t$-tests against the published scores give adjusted
    $p=0.0017$, $p<10^{-7}$, and $p=0.0196$ for CaRB~(1-1) F1, CaRB~(1-1) AUC, and
    BenchIE F1, treating the baselines as fixed references rather than as paired
    comparisons.

    On BenchIE, \diffie achieves the best average reported F1, with 9 of 10 seeds exceeding ClausIE, the strongest rule-based baseline. The gain over ClausIE is small, but \diffie substantially outperforms prior neural OpenIE systems on this stricter fact-level benchmark.

    CycleOIE is a particularly relevant comparison because our primary model is trained on a 2.5K-sentence subset of its LSOIE-examples resource. Despite using only this subset, \diffie improves over CycleOIE on CaRB F1 and on the stricter CaRB~(1-1) F1/AUC metrics, although CycleOIE retains higher standard CaRB AUC. On WiRe57, \diffie trails OpenIE6, indicating that its gains do not transfer uniformly across all evaluation settings.

\section{Ablation Study}

    All ablations are evaluated on the CaRB development set, the only one of our four benchmarks with a designated development split. BenchIE and WiRe57 provide test-only resources.

    \subsection{Effect of Training Data Size}
    \label{sec:ablations-data}

    Table~\ref{tab:data_ablation} reports CaRB and CaRB (1-1) F1 on the development set across all training configurations. The two metrics tell different stories above \textsc{lsoie-ex-1.25k}: CaRB F1 is essentially flat across the three largest curated-only configurations (52.6, 52.4, 52.5), while CaRB (1-1) F1 peaks at \textsc{lsoie-ex-2.5k} (53.1), 1.8 points above \textsc{lsoie-ex-1.25k} (51.3) and 1.7 points above \textsc{lsoie-ex-full} (51.4). We adopt \textsc{lsoie-ex-2.5k} as the primary configuration on the basis of this CaRB (1-1) advantage. Augmenting with raw LSOIE data degrades performance in both metrics, and the degradation grows with the volume of raw data added: \textsc{+lsoie-full} falls 5.6 F1 points below \textsc{lsoie-ex-full} on CaRB and 7.4 points on CaRB (1-1). The two sources differ in annotation construction and target labels, not only in size, so we read this degradation as annotation-distribution mismatch rather than as evidence that the original LSOIE labels are noisier.

    These results are consistent with the low-resource motivation of \cite{cycleoie2025}: for \diffie, more data is not automatically better. The curated \textsc{lsoie-ex-2.5k} subset outperforms both the larger curated set and the configurations augmented with raw LSOIE, suggesting that supervision quality and span alignment are more important than raw corpus size in this setting.

    \begin{table}[t]
    \centering
    \small
    \begin{tabular}{lcc}
    \toprule
    Configuration & CaRB F1 & CaRB (1-1) F1 \\
    \midrule
    \textsc{lsoie-ex-50}    & $14.4 \pm 0.2$ & $16.1 \pm 0.3$ \\
    \textsc{lsoie-ex-250}   & $39.0 \pm 0.1$ & $38.9 \pm 0.2$ \\
    \textsc{lsoie-ex-1.25k} & $52.6 \pm 0.2$ & $51.3 \pm 0.2$ \\
    \textsc{lsoie-ex-2.5k}  & $52.4 \pm 0.2$ & $53.1 \pm 0.2$ \\
    \textsc{lsoie-ex-full}  & $52.5 \pm 0.1$ & $51.4 \pm 0.2$ \\
    \midrule
    \quad + \textsc{lsoie-20k}  & $51.7 \pm 0.2$ & $49.3 \pm 0.2$ \\
    \quad + \textsc{lsoie-full} & $46.9 \pm 0.3$ & $44.0 \pm 0.2$ \\
    \bottomrule
    \end{tabular}
    \caption{CaRB and CaRB (1-1) F1 on the development set for all training
      configurations (mean $\pm$ std over 10 seeds). Upper block: curated-only
      subsets of LSOIE-examples. Lower block: \textsc{lsoie-ex-full} augmented
      with raw LSOIE data.}
    \label{tab:data_ablation}
    \end{table}

    \subsection{Test-Time Compute Scaling}
    \label{sec:ablations-nstudy}
    A central property of \diffie is that the number of denoising
samples $n$ is an inference-time hyperparameter, allowing a tunable
trade-off between extraction quality and compute. Full sensitivity tables for
$n$, the output budget $k$, and the clustering threshold $\tau$, together with
the exact-frequency extractor baseline, are reported in
Appendix~\ref{sec:appendix-sensitivity}. Performance improves with larger
sample pools and then saturates, while lenient matching consistently improves
over exact-frequency aggregation. The single CaRB-dev-selected setting
($n=512$, $k=4$, $\tau=0.9$) is used for every reported test result; it is
within 2.0 F1 of the best diagnostic sweep point on BenchIE and matches the
best point on WiRe57.

\paragraph{Inference cost.}
Table~\ref{tab:efficiency} reports end-to-end throughput and peak memory on a
single NVIDIA A100 over a CaRB development subset. Lowering $n$ from 512 to 64
costs 1.5 F1 and to 16 costs 3.0 F1, while raising throughput by
$6.6\times$ and $15\times$; clustering and ranking account for 0.2--1.0\% of
runtime. \diffie is not faster than fixed-slot tagging in absolute terms:
DetIE processes 686 sentences per second against 14.2 at $n{=}16$. The claim
is controllable inference cost, not raw speed. Against direct LLM prompting
the comparison is favorable: at $n{=}16$ \diffie matches a prompted
Qwen3-30B-A3B (49.3 vs.\ 49.5 F1) at roughly $50\times$ the throughput and a
small fraction of the memory. We also evaluate the released DetIE-LSOIE
checkpoint under our evaluator (45.1 F1). That comparison shares an evaluator
but not annotation targets---DetIE-LSOIE trains on original LSOIE token
labels, \diffie on CycleOIE annotations---so we report it as a same-evaluator
reference point rather than a controlled training comparison.

    \begin{table}[t]
    \centering
    \small
    \setlength{\tabcolsep}{3.5pt}
    \begin{tabular}{lrrrr}
    \toprule
    System & $n$ & F1 & sent/s & VRAM \\
    \midrule
    DetIE-LSOIE                & ---  & 45.1 & 686.0 & $\approx$0.7 \\
    Qwen3-30B-A3B (prompted)   & ---  & 49.5 &  0.28 & $\approx$80 \\
    \midrule
    \multirow{3}{*}{\diffie}
                               &  16  & 49.3 & 14.2  & 0.55 \\
                               &  64  & 50.8 &  6.3  & 0.65 \\
                               & 512  & 52.3 &  0.95 & 1.74 \\
    \bottomrule
    \end{tabular}
    \caption{End-to-end inference on the CaRB development set, single NVIDIA
      A100. F1 is CaRB F1, \emph{sent/s} is sentences per second, and
      \emph{VRAM} is peak GPU memory in GB. The DetIE memory figure is
      estimated; the Qwen figure is vLLM's memory reservation rather than the
      model's parameter footprint.}
    \label{tab:efficiency}
    \end{table}

    \subsection{D3PM-uniform vs.\ absorbing-state diffusion}
    \label{sec:ablations-mdlm}

    We compare D3PM-uniform against MDLM \citep{sahoo2024simple}, the standard
    absorbing-state discrete diffusion formulation, training both systems on
    \textsc{lsoie-ex-2.5k} and \textsc{lsoie-ex-full}.
    For MDLM we sweep noise schedule (cosine, linear, log-linear,
    mutual-information), number of sampling steps (8--64), temperature, and
    remasking strategy on the CaRB development set, and report the best
    configuration found for each training set.

    Results are shown in Table~\ref{tab:mdlm_ablation}.
    D3PM-uniform outperforms MDLM at both training set sizes; the $\Delta$ rows
    show the advantage is $+2.2$/$+3.4$ points at \textsc{lsoie-ex-2.5k} and
    widens to $+4.1$/$+2.9$ at \textsc{lsoie-ex-full}.
    This is consistent with \citet{schiff2025simple}, who show that uniform-noise
    diffusion can match or exceed absorbing-state diffusion in small-vocabulary
    regimes; our four-symbol tag vocabulary ($|\mathcal{V}|=4$) is an extreme
    instance of this effect.

    \begin{table}[h]
    \centering
    \small
    \begin{tabular}{llcc}
    \toprule
    Data & System & CaRB & CaRB (1-1) \\
    \midrule
    \multirow{3}{*}{\textsc{lsoie-ex-2.5k}}
      & Uniform & $52.4 \pm 0.2$ & $53.1 \pm 0.2$ \\
      & MDLM         & $50.2 \pm 0.2$ & $49.7 \pm 0.2$ \\
      & $\Delta$     & $+2.2$         & $+3.4$         \\
    \midrule
    \multirow{3}{*}{\textsc{lsoie-ex-full}}
      & Uniform & $52.5 \pm 0.1$ & $51.4 \pm 0.2$ \\
      & MDLM         & $48.4 \pm 0.3$ & $48.5 \pm 0.3$ \\
      & $\Delta$     & $+4.1$         & $+2.9$         \\
    \bottomrule
    \end{tabular}
    \caption{D3PM-uniform vs.\ MDLM on the CaRB development set. Each score entry reports mean $\pm$ std F1 over 10 seeds for CaRB and CaRB (1-1); $\Delta$ denotes D3PM-uniform minus MDLM within each data configuration.}
    \label{tab:mdlm_ablation}
    \end{table}

    \subsection{Is diffusion necessary?}
    \label{sec:ablations-mcdropout}

    Sample aggregation could in principle be driven by any stochastic tagger.
    To isolate the contribution of reverse diffusion, we train an MC-dropout
    sequence tagger that shares everything else with \diffie: the same
    training instances, the same \texttt{bert-base-uncased} encoder, the same
    B/S/R/O label space, longest-span construction, and lenient
    clustering and ranking. Full-encoder MC dropout replaces reverse diffusion
    as the candidate generator, and we select its best configuration on the
    CaRB development set from the same grid.

    The control reaches 37.5 CaRB F1 against 52.3 for \diffie
    (Table~\ref{tab:mcdropout}), and 37.3 against 53.1 under one-to-one
    matching. The gap is driven by recall: across the whole tagger sweep,
    recall never exceeds 27.3 on CaRB or 27.9 on CaRB~(1-1), whereas \diffie
    reaches 45.8 and 48.0. The control is therefore candidate-pool limited, as
    no setting of the shared clustering stage can recover triplets the tagger
    never proposes. This attributes the gain to reverse-diffusion candidate
    generation rather than to repeated sampling and clustering alone.

    \begin{table}[h]
    \centering
    \small
    \begin{tabular}{lcccc}
    \toprule
    System & F1 & P & R & AUC \\
    \midrule
    MC-dropout tagger & 37.5 & 68.9 & 25.8 & 21.3 \\
    \diffie           & 52.3 & 61.0 & 45.8 & 37.0 \\
    \bottomrule
    \end{tabular}
    \caption{Matched non-diffusion control on the CaRB development set. Both
      systems share encoder, training instances, label space, triplet
      construction, and aggregation; only the candidate generator differs.}
    \label{tab:mcdropout}
    \end{table}

\section{Discussion}
\label{sec:discussion}

\paragraph{Why diffusion fits OpenIE.}
OpenIE is a multi-output task by nature: a sentence usually contains
several valid triplets, so any single tagging is incomplete. Diffusion
fits this well because its reverse process is stochastic — running it
several times on the same input gives different valid outputs, which
matches how OpenIE ground truth is structured. Other non-autoregressive
methods handle this by building multiplicity into the architecture
(fixed slots, bipartite matching) or the decoder (autoregressive
iteration); \diffie instead lets sample diversity do the work, and
turns the candidate-pool size into an inference-time knob. That knob is a
cost control: \diffie is slower than fixed-slot tagging but substantially
faster and more memory-efficient than direct LLM prompting. Its sample count
allows quality to be traded against throughput after training
(Table~\ref{tab:efficiency}).
Table~\ref{tab:appendix-n-extractor} shows that this pool converges,
and that lenient-match beats frequency aggregation across
all $n$ shows that the candidate pool contains many near-equivalent
triplets that differ only in span boundaries — variation that
clustering recovers but exact-match aggregation fragments.
The gains show up most on CaRB~(1-1) and BenchIE, the two benchmarks
built to penalize over-extraction; on WiRe57, whose references prefer
short, syntactically tight extractions, \diffie trails OpenIE6. We
attribute this to the style of our LSOIE-derived training supervision,
which by construction rewards exhaustive extraction and multi-token
relation spans that include modifiers and determiners — a style well
aligned with CaRB but at odds with WiRe57's curated references.

\paragraph{Uniform vs.\ absorbing-state diffusion.}
Uniform-noise diffusion outperforms MDLM at both training sizes, and
the gap widens with more data (\S\ref{sec:ablations-mdlm}). We
attribute this to a structural difference in the reverse process:
under absorbing-state corruption, each position transitions once from
\texttt{[MASK]} to a tag and stays there, so an early wrong commitment
is locked in unless explicitly remasked at inference; under uniform
corruption, positions can transition between non-mask tags at any
step, letting the denoiser revise earlier decisions as more context
becomes available. This matters for OpenIE because role tags are
jointly constrained. Our MDLM sweep included remasking strategies and
uniform diffusion still won, consistent with this account. Together
with \citet{schiff2025simple}, this suggests uniform corruption is the
better default for structured prediction over small label vocabularies.

\section{Conclusion}
We introduced \diffie, which treats Open Information Extraction as
conditional discrete diffusion over per-token role tags and uses the
stochasticity of the reverse process as the extraction mechanism
itself. Independent denoising trajectories conditioned on the same
sentence yield a diverse pool of candidate triplets, which a
lenient-match extractor clusters and ranks to recover multiple valid
extractions without autoregressive decoding or train-time fixed slots.
Because both the candidate-pool size and the number of returned
clusters are inference-time choices, \diffie exposes test-time compute
as a tunable quality--cost axis. Across four OpenIE benchmarks,
\diffie achieves the best reported CaRB~(1-1), both F1 and AUC, and beats the
strongest rule-based system on BenchIE while substantially
outperforming all prior neural OpenIE systems with reported BenchIE
numbers. It remains competitive on standard CaRB and WiRe57, and has the
best average score among systems that report all four benchmarks. A matched
MC-dropout tagger sharing our encoder, labels, and aggregation does not
reproduce these gains, which places them in candidate generation. We
further find that uniform discrete diffusion outperforms
absorbing-state diffusion in this four-tag setting, which we attribute
to the reverse process's ability to revise earlier decisions, and
that supervision quality matters more than corpus size for
sample-aggregation models. More broadly, our results indicate that
diffusion stochasticity is a useful mechanism for structured
prediction tasks with multiple valid outputs, and we leave its
application to other multi-reference NLP tasks to future work.

\section*{Acknowledgments}

This work was carried out within the state assignment under the research theme
``Methods and technologies of a cloud-based, service-oriented digital platform
for collecting, storing, and processing large volumes of multi-format
interdisciplinary data and knowledge, based on the use of artificial
intelligence, component-based and model-driven approaches, and machine
learning'' (code FWEW-2026-0012, State Registration No. 126021217141-8).

\section*{Limitations}

\paragraph{Longest-span construction heuristic.}
We construct triplets from denoised tag sequences by taking the longest
contiguous run of each role tag (\S\ref{sec:method}). This filters
isolated tag errors but discards information when a role is
discontinuous in the prediction. On the CaRB dev set, 28.9\% of
denoised samples contain at least one discontinuous role-tag run, most
often in Relations (18.6\%) and Objects (12.2\%), discarding 2.4
tokens on average when triggered — roughly 0.7 tokens per sample
unconditionally. Because \diffie aggregates $n=512$ trajectories per
sentence and ranks clusters by mass, these occasional drops are
absorbed in aggregation rather than directly producing wrong
extractions — the same fact is typically recovered by other
trajectories whose longest-span construction succeeds. Alternative
construction rules that preserve multiple maximal spans per role are
straightforward to plug in post-sampling and may further improve
recall; we leave this to future work. More generally, the four-tag scheme
assigns each token a single role within a trajectory, so overlapping roles
and discontinuous arguments cannot be represented inside one sample.
Sampling removes the cap on how many triplets a sentence yields, but not
this per-trajectory restriction.

\paragraph{Training data scope.}
All experiments use English supervision from CycleOIE's curated
\textsc{lsoie-examples} \citep{cycleoie2025}. We do not
evaluate on other languages or out-of-domain corpora. Our
span-alignment filter additionally discards roughly 40\% of instances
whose GPT-generated arguments do not match source spans verbatim;
training on datasets with cleaner extractive spans could increase
usable supervision. We have not carried out a stratified audit of the
discarded instances, so we cannot say whether the filter correlates with
relation type, argument length, discontinuity, or syntactic construction;
our conclusions are scoped to the retained, extractive portion of the data.

\balance
\bibliography{custom}

\clearpage
\appendix

\section{Full Results with Standard Deviations}
\label{sec:appendix-full-results}

Tables~\ref{tab:full_results_carb_group}
and~\ref{tab:full_results_benchie_wire57_group} report precision,
recall, F1, and AUC where applicable for \diffie on CaRB, CaRB (1-1),
BenchIE, and WiRe57 across 10 samples. All metrics are shown in percentage
points. Each panel reports the mean and standard deviation across samples
in its final row.

\begin{table}[H]
\centering
\scriptsize
\setlength{\tabcolsep}{1.5pt}
\begin{tabular}{l*{10}{c}}
\toprule
\multicolumn{11}{c}{\textbf{CaRB}} \\
\midrule
Metric & \multicolumn{10}{c}{Sample} \\
\cmidrule(lr){2-11}
& 0 & 1 & 2 & 3 & 4 & 5 & 6 & 7 & 8 & 9 \\
\midrule
AUC & 37.4 & 36.9 & 37.2 & 37.1 & 37.2 & 37.0 & 36.9 & 37.0 & 37.2 & 37.3 \\
P   & 63.2 & 63.1 & 65.7 & 66.2 & 62.5 & 64.5 & 63.2 & 64.9 & 62.6 & 65.1 \\
R   & 44.8 & 44.5 & 43.1 & 43.4 & 44.7 & 44.0 & 44.3 & 43.7 & 44.8 & 43.8 \\
F1  & 52.4 & 52.2 & 52.1 & 52.4 & 52.1 & 52.3 & 52.1 & 52.2 & 52.2 & 52.4 \\
\midrule
Mean $\pm$ Std
& \multicolumn{10}{c}{$\mathrm{AUC}=37.1 \pm 0.2,\ P=64.1 \pm 1.3$} \\
& \multicolumn{10}{c}{$R=44.1 \pm 0.6,\ F1=52.2 \pm 0.1$} \\
\bottomrule
\end{tabular}

\vspace{0.75em}

\begin{tabular}{l*{10}{c}}
\toprule
\multicolumn{11}{c}{\textbf{CaRB (1-1)}} \\
\midrule
Metric & \multicolumn{10}{c}{Sample} \\
\cmidrule(lr){2-11}
& 0 & 1 & 2 & 3 & 4 & 5 & 6 & 7 & 8 & 9 \\
\midrule
AUC & 34.6 & 34.3 & 34.4 & 34.7 & 34.5 & 34.4 & 34.4 & 34.4 & 34.2 & 34.6 \\
P   & 59.8 & 62.1 & 58.9 & 60.9 & 59.8 & 59.9 & 60.8 & 60.3 & 60.3 & 59.4 \\
R   & 46.1 & 44.3 & 46.1 & 46.0 & 45.8 & 45.7 & 45.3 & 45.4 & 45.1 & 45.9 \\
F1  & 52.1 & 51.7 & 51.7 & 52.4 & 51.8 & 51.8 & 51.9 & 51.8 & 51.6 & 51.8 \\
\midrule
Mean $\pm$ Std
& \multicolumn{10}{c}{$\mathrm{AUC}=34.5 \pm 0.1,\ P=60.2 \pm 0.9$} \\
& \multicolumn{10}{c}{$R=45.6 \pm 0.5,\ F1=51.9 \pm 0.2$} \\
\bottomrule
\end{tabular}
\caption{\diffie results on CaRB and CaRB (1-1) for the \textsc{lsoie-ex-2.5k} configuration.}
\label{tab:full_results_carb_group}
\end{table}

\begin{table}[H]
\centering
\scriptsize
\setlength{\tabcolsep}{1.5pt}
\begin{tabular}{l*{10}{c}}
\toprule
\multicolumn{11}{c}{\textbf{BenchIE}} \\
\midrule
Metric & \multicolumn{10}{c}{Sample} \\
\cmidrule(lr){2-11}
& 0 & 1 & 2 & 3 & 4 & 5 & 6 & 7 & 8 & 9 \\
\midrule
P  & 38.6 & 38.5 & 37.9 & 39.0 & 38.3 & 38.2 & 38.6 & 37.5 & 38.5 & 38.9 \\
R  & 31.0 & 30.7 & 31.0 & 31.3 & 31.0 & 31.0 & 31.3 & 30.4 & 30.9 & 31.3 \\
F1 & 34.4 & 34.2 & 34.1 & 34.7 & 34.2 & 34.2 & 34.6 & 33.6 & 34.3 & 34.7 \\
\midrule
Mean $\pm$ Std
& \multicolumn{10}{c}{$P=38.4 \pm 0.4,\ R=31.0 \pm 0.3,\ F1=34.3 \pm 0.3$} \\
\bottomrule
\end{tabular}

\vspace{0.75em}

\begin{tabular}{l*{10}{c}}
\toprule
\multicolumn{11}{c}{\textbf{WiRe57}} \\
\midrule
Metric & \multicolumn{10}{c}{Sample} \\
\cmidrule(lr){2-11}
& 0 & 1 & 2 & 3 & 4 & 5 & 6 & 7 & 8 & 9 \\
\midrule
P  & 42.2 & 42.1 & 44.1 & 43.1 & 42.9 & 43.9 & 42.7 & 43.5 & 43.0 & 42.7 \\
R  & 30.4 & 30.7 & 31.3 & 30.8 & 30.5 & 32.0 & 31.5 & 31.0 & 31.1 & 31.7 \\
F1 & 35.4 & 35.5 & 36.6 & 35.9 & 35.6 & 37.0 & 36.3 & 36.2 & 36.1 & 36.4 \\
\midrule
Mean $\pm$ Std
& \multicolumn{10}{c}{$P=43.0 \pm 0.6,\ R=31.1 \pm 0.5,\ F1=36.1 \pm 0.5$} \\
\bottomrule
\end{tabular}
\caption{\diffie results on BenchIE and WiRe57 for the \textsc{lsoie-ex-2.5k} configuration.}
\label{tab:full_results_benchie_wire57_group}
\end{table}

\section{Sensitivity Analyses}
\label{sec:appendix-sensitivity}

Tables~\ref{tab:appendix-sensitivity} and~\ref{tab:appendix-n-extractor}
vary $\tau$, $k$, and $n$ one at a time, holding the other two at the
CaRB-dev-selected values. BenchIE and WiRe57 have no development split, so
their sweeps are diagnostic only and do not determine any reported test
setting. The selected operating point is $\tau=0.9$, $k=4$, and $n=512$.

\begin{table}[H]
\centering
\small
\setlength{\tabcolsep}{5pt}
\begin{tabular}{llrrr}
\toprule
Sweep & Value & CaRB dev & BenchIE & WiRe57 \\
\midrule
\multirow{6}{*}{$\tau$}
 & 0.50 & 43.9 & 22.2 & 18.6 \\
 & 0.60 & 44.8 & 22.9 & 19.6 \\
 & 0.70 & 47.3 & 25.0 & 25.7 \\
 & 0.80 & 51.7 & 28.8 & 30.5 \\
 & \textbf{0.90} & \textbf{52.3} & \textbf{34.3} & \textbf{36.1} \\
 & 0.95 & 49.8 & 36.3 & 34.8 \\
\midrule
\multirow{6}{*}{$k$}
 & 1  & 43.4 & 20.0 & 17.4 \\
 & 2  & 51.8 & 30.2 & 27.0 \\
 & \textbf{4}  & \textbf{52.3} & \textbf{34.3} & \textbf{36.1} \\
 & 6  & 52.2 & 31.9 & 36.1 \\
 & 8  & 52.2 & 29.4 & 34.4 \\
 & 10 & 52.2 & 27.1 & 32.4 \\
\midrule
\bottomrule
\end{tabular}
\caption{One-at-a-time sensitivity of F1 (percentage points) to the
  clustering threshold $\tau$ and output budget $k$. Bold rows mark the
  CaRB-dev-selected values used for all reported results.}
\label{tab:appendix-sensitivity}
\end{table}

Table~\ref{tab:appendix-n-extractor} reports sample-count scaling across all
benchmarks and additionally compares the lenient-match extractor with
exact-frequency aggregation on CaRB development data. The latter groups
triplets only when all three span boundaries match exactly, fragmenting
posterior mass across boundary-shifted realizations of the same fact. Lenient
matching is consistently stronger once multiple samples are available, and
all benchmarks show diminishing returns at larger $n$.

\begin{table}[H]
\centering
\scriptsize
\setlength{\tabcolsep}{2.5pt}
\begin{tabular}{rrrrrrr}
\toprule
& \multicolumn{2}{c}{CaRB} & \multicolumn{2}{c}{CaRB (1-1)}
& BenchIE & WiRe57 \\
\cmidrule(lr){2-3}\cmidrule(lr){4-5}\cmidrule(lr){6-6}\cmidrule(lr){7-7}
$n$ & Len. & Freq. & Len. & Freq. & Len. & Len. \\
\midrule
1    & 39.8 & 39.8 & 40.7 & 40.7 & 13.7 & 15.3 \\
2    & 46.2 & 45.9 & 47.0 & 46.8 & 19.9 & 19.7 \\
4    & 48.3 & 47.3 & 49.7 & 49.1 & 25.8 & 29.2 \\
8    & 48.3 & 46.6 & 50.0 & 48.7 & 28.2 & 30.2 \\
16   & 50.6 & 48.1 & 51.2 & 49.0 & 30.5 & 32.8 \\
32   & 50.6 & 48.8 & 51.6 & 49.6 & 31.9 & 33.2 \\
64   & 51.5 & 49.1 & 52.1 & 50.0 & 33.9 & 34.3 \\
128  & 51.8 & 49.4 & 52.5 & 50.2 & 34.7 & 35.7 \\
256  & 52.6 & 49.3 & 53.1 & 50.1 & 35.1 & 36.0 \\
\textbf{512} & \textbf{52.3} & 49.1 & \textbf{53.1} & 50.2
      & \textbf{34.3} & \textbf{36.1} \\
1024 & 52.5 & 49.1 & 52.9 & 50.0 & --- & --- \\
\bottomrule
\end{tabular}
\caption{F1 versus the number of sampled trajectories. CaRB development
  results compare lenient-match (Len.) and exact-frequency (Freq.)
  aggregation; BenchIE and WiRe57 use lenient matching. Bold entries mark the
  globally selected $n$.}
\label{tab:appendix-n-extractor}
\end{table}

\end{document}